# Determining Mosaic Resilience in Sugarcane Plants using Hyperspectral Images

Ali Zia, Jun Zhou and Muyiwa Olayemi

## Abstract

Sugarcane mosaic disease poses a serious threat to the Australian sugarcane industry, leading to yield losses of up to 30% in susceptible varieties. Existing manual inspection methods for detecting mosaic resilience are inefficient and impractical for large-scale application. This study introduces a novel approach using hyperspectral imaging and machine learning to detect mosaic resilience by leveraging global feature representation from local spectral patches. Hyperspectral data were collected from eight sugarcane varieties under controlled and field conditions. Local spectral patches were analyzed to capture spatial and spectral variations, which were then aggregated into global feature representations using a ResNet18 deep learning architecture. While classical methods like Support Vector Machines struggled to utilize spatial-spectral relationships effectively, the deep learning model achieved high classification accuracy, demonstrating its capacity to identify mosaic resilience from fine-grained hyperspectral data. This approach enhances early detection capabilities, enabling more efficient management of susceptible strains and contributing to sustainable sugarcane production.

## 1. Introduction

Sugarcane is Australia's major source of raw sugar. The sugarcane industry in Queensland accounts for approximately AUD 4 billion in value, which supports more than 22,000 jobs and 10,000 businesses1. One of the challenges sugarcane farmers faces is sugarcane disease prevention and management.

Sugarcane mosaic is a disease that has been found in all regions of the Australian sugarcane industry but is currently restricted to the Southern region particularly the Bundaberg/ Childers district. Mosaic is widespread in the Childers district, but management strategies have restricted the intensity of infestations. Mosaic can cause losses of 30% or more in susceptible varieties. Sporadic outbreaks of mosaic occur when conditions are favourable for the aphid vectors and susceptible varieties are grown.

Mosaic can disrupt disease-free (approved) seed schemes. Careful inspection and rouging of infected plants are required to maintain the integrity of disease-free seed plots.

In Australia, we have only one strain of the sugarcane mosaic virus. Sugarcane mosaic is caused by a range of different viruses in other countries. In Asia and North America, the mosaic disease is very widespread and causes small losses but over vast areas of sugarcane.

Sugarcane mosaic is managed by resistant varieties and disease-free (approved) seed. Only a small percentage of varieties are considered too susceptible to be grown commercially in Australia because we have a mild strain of the virus.

Regularly obtaining disease-free (approved) seed is important in the management of mosaic. Disease-free seed plots must be regularly inspected for mosaic and, if infected plants are found, they must be immediately destroyed to prevent further spread. Disease-free seed plots should be located away from known sources of mosaic infection.

Controlling weeds on which the aphid vectors can breed within fields and surrounding headlands may help to reduce spread. Quarantine is important to prevent the spread of infected plants to districts where the mosaic is not present and to prevent exotic strains of mosaic from entering Australia.

Finding low resilience strains early and eliminating them promptly is an important step to stop the spread of the disease. Current only manual visual inspection is used to find out these stains. Considering the size of the sugarcane field this is an impractical and time-consuming job, with the good amount of change that some plants will still be missed. In addition to that, it also depends on the expertise and bias of a human inspector.

Computer vision and machine learning have been used for the automatic detection of plant diseases. Early research captured grayscale or colour images and used traditional machine learning methods such as Support Vector Machine (SVM) and K-Nearest Neighbors (KNN) [1]–[3] as the detectors. With the success of deep learning, convolutional neural networks (CNNs) have more recently been adopted for plant disease detection [4], [5]. They can be integrated with traditional classifiers [6] to achieve improved results.

Visual symptoms of the mosaic may take some time to develop, by which time the disease epidemic may be well established. Mosaic resilence detection at the early stages of breeding programs offers a faster decision to discard susceptible lines. However, this is a challenging task because the only visible symptom that appears are mosaic patterns, which are not necessarily observable at an early stage. This motivates us to explore the information beyond the visible spectrum to gain a stronger capability of early detection. Therefore, we leverage hyperspectral imaging technology to capture data at a fine spectral resolution in both visible and near-infrared spectra. Unlike RGB images which contain three colour channels, hyperspectral images typically contain tens or hundreds of bands (channels). Fig. 1, shows rich spectral information present in hyperspectral images of sugarcane leaf, which can be extremely beneficial for early detection.

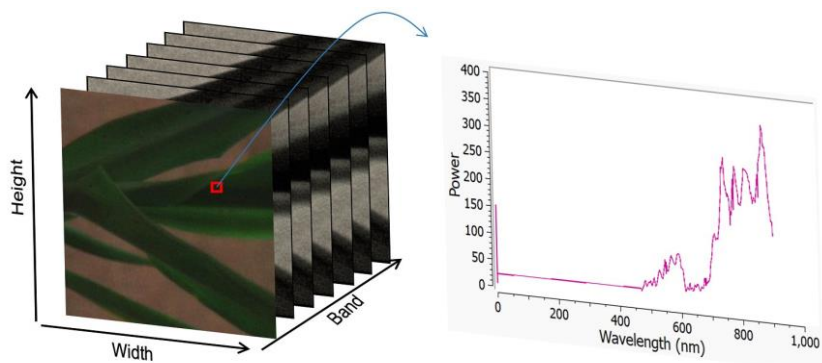

Fig. 1: An example of a hyperspectral sugarcane leaf image cube, on the right is the spectral curve of a sugarcane pixel.

With the development of imaging technologies, hyperspectral images have been increasingly used in plant disease detection [3], [7], [8]. Moghadam et al. [3] extracted features from vegetation indices, full-spectrum, and probabilistic topic models on hyperspectral images to detect tomato spotted wilt virus in capsicum plants. Xie et al. [2] detected the early-stage symptoms of grey mould disease on tomato leaves one day after the inoculation. Rumpf et al. [9] built an automatic approach for the early detection of three sugar beet diseases based on the classification of spectral reflectance data using support vector machines. However, to our knowledge, there has not been reported work on the early detection of mosaic resilience of hyperspectral imaging.

## 2. Objectives

The main objectives of this work are as follows:
- To see how useful the near-infrared range in hyperspectral imaging is to mosaic resilience detection.
- Find feasibility and accuracy of automatic classification using machine learning approaches on hyperspectral data.
- Analyse the findings and propose a plan.

## 3. Mosaic Resistance Rating and Hyperspectral Dataset

For this work, we investigated eight varieties of sugarcane plants with seven distinct rating classes. Detail of each species rating is given in Table 1. The lower the rating number the more resilient the plant is to the mosaic disease.

| Variety | Mosaic rating |
|---|---|
| CP29-116 | 6 |
| Pindar | 2 |
| Q44 | 8 |
| Q68 | 5 |
| Q78 | 9 |
| Q82 | 1 |
| 155 | 6 |
| Q205 | 7 |

Table 1: Mosaic disease resilience rating with respect to sugarcane plant species

The hyperspectral dataset was built from an indoor and outdoor environment. From each variety mentioned in the table, we collected six to nine images per class from each indoor and outdoor environment. Each variety has images with varying amounts of disease symptoms present, ranging from none to severe.

The images were captured using Ximea hyperspectral camera. This camera covers the spectral range approximately from 690 nm to 840 nm with 11 bands. The size of captured images is 216x409x11, with the first two dimensions indexing the spatial locations of pixels, and the third dimension being the spectral bands.

Before image capture, a dark reference image was captured by covering the lens of the camera with a cap. The dark reference image provides information on system noises. The plant image acquisition was done directly under the sunlight with a white spectral calibration board, also called spectralon, placed next to the plant. The white calibration board reflects ≥ 99% of the light from 400 nm to 1500 nm. It is used to ground-truth the lighting conditions.

Hyperspectral image calibration aims to remove the noise and reduce the influence caused by different lighting conditions. It includes dark calibration and white calibration. The dark calibration step subtracts the dark reference image from the original image so that the system noise of the camera can be removed.

The white calibration requires the extraction of the white calibration board from the image. This was done by a K-means clustering method to retrieve the brightest cluster in the image. After that, the top 1% brightest pixels were removed to avoid saturation. The calibration process then calculated the mean spectral curve of the remaining top 1000 brightest pixels, which were subsequently used as the reference curve of the lighting condition for the captured sugarcane image. All pixels in the image were then divided by the mean spectral curve band-by-band.

## 4. Data Processing and Machine Learning Approaches

In this section, we will our the approach and methodology that we used for determining automatic mosaic resilience in the sugarcane plant. With the sugarcane hyperspectral image dataset collected, we treat the rating of mosaic disease as a multi-classification problem, with seven classes. Before we apply the machine learning algorithm we need to preprocess the data and figure out how can be useful.

### 4.1. Data preprocessing

This step aims to reduce the influence of background so model learning and prediction are only performed on sugarcane plants. To reach this goal, the sugarcane plant was segmented from its background in the hyperspectral image. Here we treated the task as a semantic segmentation process and adopted a supervised learning approach. We first manually annotated several images so that the sugarcane plant was marked as the foreground and the rest of the image was treated as the background. The labelled images were then used to train an SVM classifier [11]. The training was done at the pixel level with the spectral curve and foreground-background labels as the input. Afterwards, the trained SVM was used to generate the semantic labels of the rest of the images to automatically segment sugarcane from the background. Fig. 2 shows an example input image, the generated semantic labels, and the segmented sugarcane image, indicating that the sugarcane segmentation is complete and highly accurate.

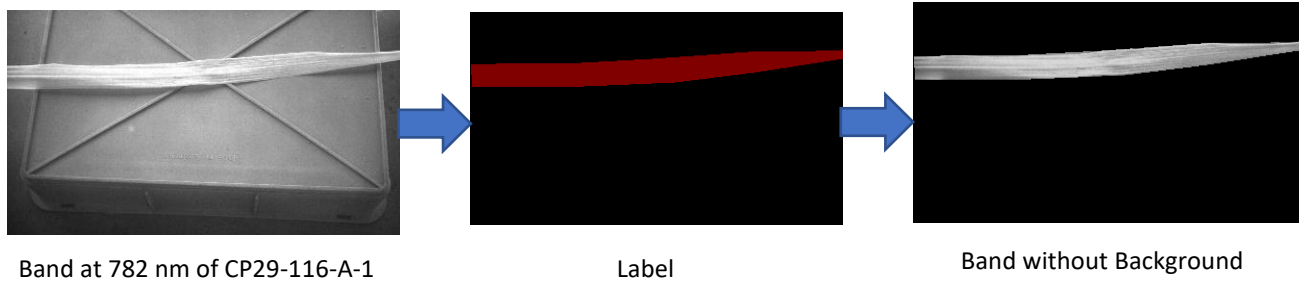

Band at 782 nm of CP29-116-A-1    Label    Band without Background

Fig. 2: Separating sugarcane from the background

### 4.2. Initial Analysis

To analyse how spectral responses can contribute to different mosaic patterns on varied species of sugarcane plants we used some analytical spectral techniques. These techniques gave us the confidence that hyperspectral images are contributing more than simple RGB images. In this report, we will present two of these techniques that contributed to our decision to move forward with hyperspectral images.

Firstly, we use the mean spectral curve to observe changes in different varieties. This gives us the baseline confidence that the two varieties are different spectrally. The mean spectral curve is defined as the mean value of all the pixels of the leaf in each spectral band. Fig 3. Shows mean spectral curve of hyperspectral images of Pindar and Q78 variety. The circular areas highlight some of the changes that are present in the curve with respect to each other.

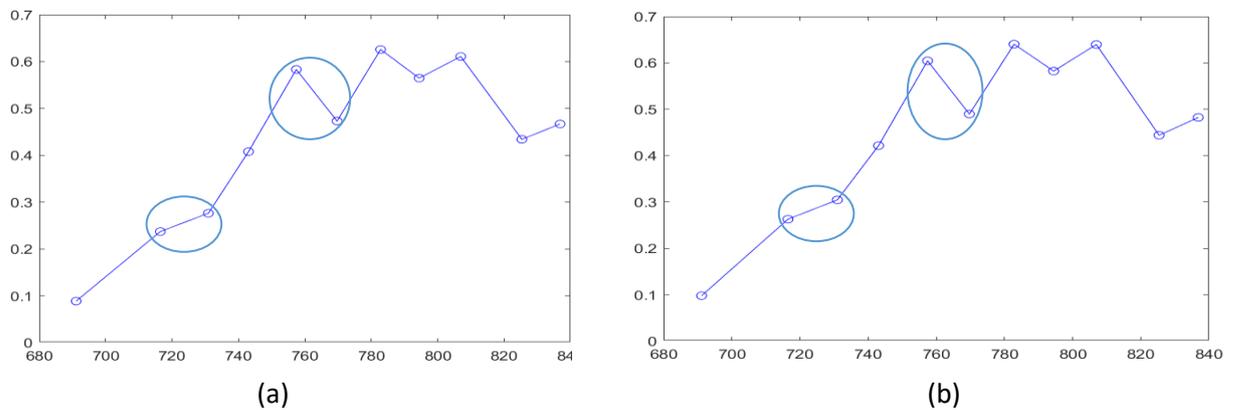

(a)    (b)

Fig. 3: Mean spectral curve of hyperspectral images of (a) Pindar and (b) Q78 variety

Next, to analyse mosaic patterns more closely in different varieties of sugarcane plants we have to perform neighbourhood spectral analysis. For this, we used the graph Laplacian based on Spectral Angular Mapper (SAM) distance. Laplacian matrix **L** is obtained by establishing a graph Laplacian of *3 × 3* neighbourhood pixels. The Laplacian matrix **L** is calculated as **L = D − W**, where D is a diagonal matrix **D(i, i ) = $\Sigma_j$ W(i, j )** and **W** is a symmetric matrix whose off-diagonal entries are spectral affinities calculated by SAM. Given two n-dimensional spectral vectors x and y, their spectral angle is defined as follows:

$$\theta(x,y) = cos^{-1}\left(\frac{\sum_{i=1}^{n} x_i y_i}{\left(\sum_{i=1}^{n} x_i^2\right)^{\frac{1}{2}} * \left(\sum_{i=1}^{n} y_i^2\right)^{\frac{1}{2}}}\right)$$

where n is the number of band images. Fig. 4 shows the result of the apply this method to Pindar and Q78 images. Here we can see different mosaic patterns highlighted in each image. This means that each variety develops a slightly different pattern than the other one. Now by using machine learning we will try to identify these patterns by classifying each variety.

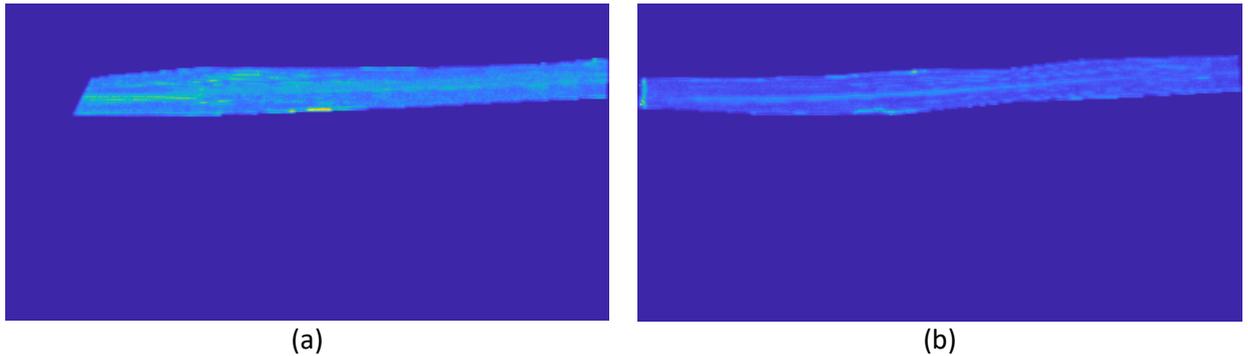

(a)                      (b)

Fig. 4: Different Spectral results after applying graph Laplacian with SAM on (a) Pindar and (b) Q78 hyperspectral images

### 4.3. Classic Machine Learning approaches

Initially, we used the Support Vector Machine(SVM) approach to do regression and classification of the data. If the kernel used in SVM performs the classification task then it is called Support Vector Classifier (SVC) and if it performs a regression then it is called Support Vector Regressor (SVR).

SVMs are so-called maximum-margin classifiers. This means that they will attempt to maximize the distance between the closest vectors of each class and the line. These closest vectors are called support vectors, and hence the name Support Vector Machine.

Using SVC and SVR we did not get very good results, but they were still much better than random results. Details discussion about the results and shortfalls of these methods are discussed in the "Results and Discussion" section.

### 4.4. Deep Learning approach

We leverage the recent progress on deep learning and adopts a ResNet [10] as the preferred algorithm for this task. The ResNet architecture consists of multiple CNN blocks with the residual mechanism or called residual blocks.

To solve a complex problem, we sometimes stack some additional layers in the Deep Neural Networks which improves the results in terms of accuracy and performance. The intuition behind adding more layers is that these layers progressively learn more complex features. For example, in the case of recognising images, the first layer may learn to detect edges, the second layer may learn to identify textures and similarly the third layer can learn to detect objects and so on. But it has been found that there is a maximum threshold for depth with the traditional Convolutional neural network model. Adding more layers on top of a network can also degrade performance.

This problem of training very deep networks has been alleviated with the introduction of ResNet or residual networks and these Resnets are made up from Residual Blocks. As shown in Fig. 5 in ResNet, there is a direct connection that skips some layers(may vary in different models) in between. This connection is called 'skip connection' and is the core of residual blocks. Due to this skip connection, the output of the layer is not the same now. Without using this skip connection, the input gets multiplied by the weights of the layer followed by adding a bias term.

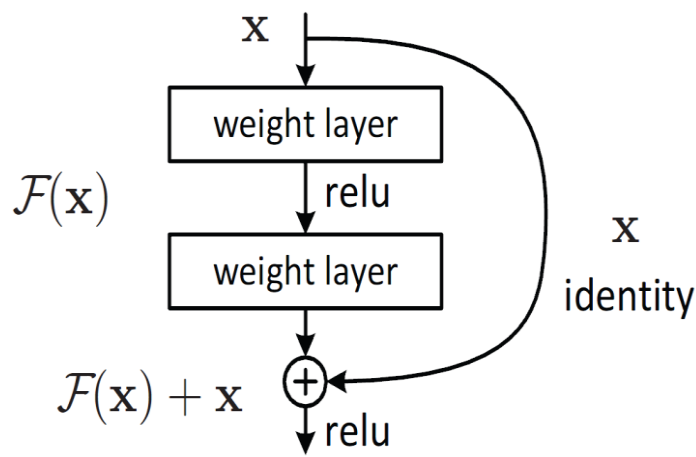

Fig. 5: Residual block used in ResNet

The skip connections in ResNet solve the problem of vanishing gradient in deep neural networks by allowing this alternate shortcut path for the gradient to flow through. The other way that these connections help is by allowing the model to learn the identity functions which ensures that the higher layer will perform at least as good as the lower layer, and not worse. Let me explain this further.

For our data, we use ResNet18, which is 18 layers deep. Details related to how data was formulated, results and discussions are presented in the next section.

## 5. Results and Discussion

In total 58 indoor and 53 outdoor images were captured plants. However, this is insufficient to train a good deep learning model. Furthermore, the spatial size of the original image, on which leaf pixels are present is of a different shape of each image, which makes training image by image infeasible. To address this problem, we used a sliding window to travel across each image and extracted *n×n×11* image cubes. Where *n* is the number of neighbourhood pixels. The image cubes were saved if the sugarcane foreground pixels occupied all the pixels. Each image cube inherited the label from the parent image, i.e. resilience rating number. This data was then split into training, validation, and test sets with the ratio of 6:2:2.

For SVM data was used by stretching out the spatial resolution and appending spectral data to create an instance. For example, if we have a patch size of 19x19x11. We have instance of 361x11 => 1x3971. For a variety of SVCs and SCRs, we got accuracy ranging from approximately 31% to 38% for our data. Given seven classes it is still better than the random result. The low accuracy was because when data was stretched out its neighbouring spatial and spectral information was lost. Results captured in the initial analysis (by using graph Laplacian and SAM) were obtained by using neighbouring information which is not exploited using these methods. Also, the classical method does not exploit all information present in hyperspectral images effectively. Because of these reasons, we looked for a machine learning method that can exploit spatial-spectral information in a better way.

For ResNet, we used data augmentation to increase the number of patches and make data prune to rotation and translation transformation. Patches were randomly translated up to three pixels horizontally and vertically and rotate the images with an angle up to 20 degrees. Fig. 6 shows the training accuracy and loss curves for 19x19x11 patch size. The curves show that the model has converged. We get similar curves for other patch sizes as well.

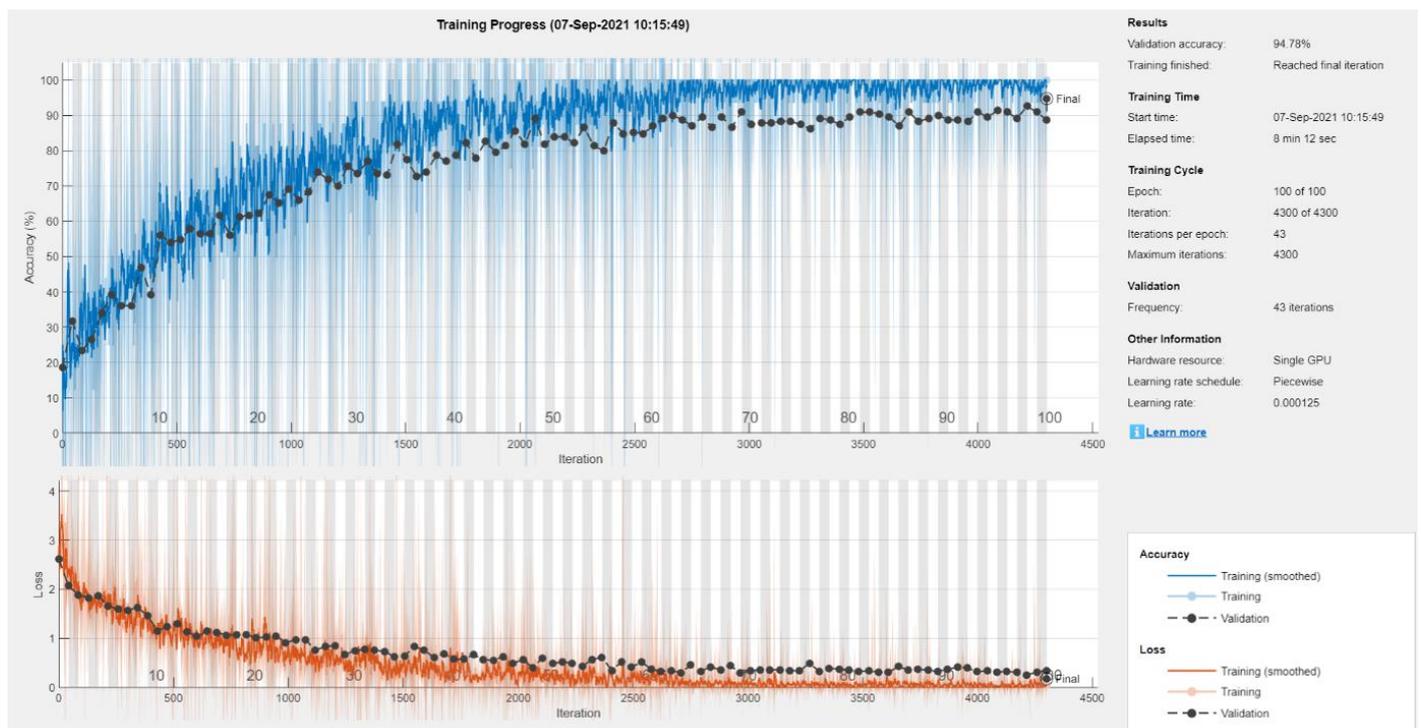

Fig. 6: ResNet training accuracy and loss curves

Table 2 shows the results obtained by using different patch sizes and strides in the sliding window operation. Stride is defined as after how many pixels the patch is going to be select in sliding window operation. For example, the last two rows have no overlap because neighbourhood size is 15 and stride is 15 as well. By observing the table we can conclude that if we have a decent number of patches our accuracy becomes quite good and if the number of patches becomes small it decreases. However, it also needs to be noted that overlapping also affects the result too, more overlapping gives us a slightly better result.

| Patch Size | Patch creation Stride | No. of training patches | No. of Validation \| Testing patches | Place | Training Error | Validation Accuracy | Test Accuracy |
| --- | --- | --- | --- | --- | --- | --- | --- |
| 19x19x11 | 9 | 703 | 230 \| 234 | Indoor | 0% | 94.7826% | 98.2906% |
| 19x19x11 | 9 | 630 | 207 \| 209 | Outdoor | 2.6984% | 93.7198% | 96.1722% |
| 17x17x11 | 9 | 930 | 306 \| 309 | Indoor | 0.32258% | 95.4248% | 92.8803% |
| 17x17x11 | 9 | 747 | 246 \| 249 | Outdoor | 2.4096% | 91.4634% | 95.1807% |
| 15x15x11 | 9 | 1238 | 408 \| 412 | Indoor | 2.4233% | 92.6471% | 91.2621% |
| 15x15x11 | 9 | 881 | 289 \| 292 | Outdoor | 3.4052% | 92.0415% | 91.4384% |
| 15x15x11 | 15 | 458 | 149 \| 152 | Indoor | 3.0568% | 79.8658% | 85.5263% |
| 15x15x11 | 15 | 328 | 104 \| 109 | Outdoor | 30.4878% | 67.3077% | 62.3853% |

Table 2: Results Obtained from Resnet

## 6. Conclusion and findings

In this work, we observed that the near-infrared range in hyperspectral imaging is beneficial in determining mosaic. Also, using a deep neural network with hyperspectral data can be used to create an automatic classification of mosaic resilience detection for sugarcane plants. We also found during experiments that when there is more data (patches) accuracy variation is less (approximately 3-4%). However, if patches are relatively less then it can go up to approximately 8%. Moreover, higher spatial/spectral resolution increase the confidence measure.

## REFERENCES

[1] K. R. Gavhale and U. Gawande, "An overview of the research on plant leaves disease detection using image processing techniques," *IOSR Journal of Computer Engineering*, vol. 16, no. 1, pp. 10–16, 2014.

[2] C. Xie, C. Yang, and Y. He, "Hyperspectral imaging for classification of healthy and gray mold diseased tomato leaves with different infection severities," *Computers and Electronics in Agriculture*, vol. 135, pp. 154–162, 2017.

[3] P. Moghadam, D. Ward, E. Goan, S. Jayawardena, P. Sikka, and E. Hernandez, "Plant disease detection using hyperspectral imaging," in *2017 International Conference on Digital Image Computing: Techniques and Applications*, 2017, pp. 1–8.

[4] K. P. Ferentinos, "Deep learning models for plant disease detection and diagnosis," *Computers and Electronics in Agriculture*, vol. 145, pp. 311–318, 2018.

[5] S. P. Mohanty, D. P. Hughes, and M. Salathe, "Using deep learning for´ image-based plant disease detection," *Frontiers in Plant Science*, vol. 7, p. 1419, 2016.

[6] M. Turköglu and D. Hanbay, "Plant disease and pest detection using deep learning-based features," *Turkish Journal of Electrical Engineering & Computer Science*, vol. 27, pp. 1636–1651, 05 2019.

[7] G. Polder, P. M. Blok, H. A. C. de Villiers, J. M. van der Wolf, and J. Kamp, "Potato virus Y detection in seed potatoes using deep learning on hyperspectral images," *Frontiers in Plant Science*, vol. 10, p. 209, 2019.

[8] K. Nagasubramanian, S. Jones, A. K. Singh, A. Singh, B. Ganapathysubramanian, and S. Sarkar, "Explaining hyperspectral imaging based plant disease identification: 3d CNN and saliency maps," *CoRR*, vol. abs/1804.08831, 2018.

[9] T. Rumpf, A.-K. Mahlein, U. Steiner, E.-C. Oerke, H.-W. Dehne, and L. Plumer, "Early detection and classification of plant diseases with support vector machines based on hyperspectral reflectance," *Computers and Electronics in Agriculture*, vol. 74, no. 1, pp. 91–99, 2010.

[10] K. He, X. Zhang, S. Ren, and J. Sun, "Deep residual learning for image recognition," in *Proceedings of the IEEE Conference on Computer Vision and Pattern Recognition*, June 2016.

[11] C. Cortes and V. Vapnik, "Support vector networks," *Machine Learning*, vol. 20, pp. 273–297, 1995.

[12] A. Vaswani, N. Shazeer, N. Parmar, J. Uszkoreit, L. Jones, A. N. Gomez, Ł. Kaiser, and I. Polosukhin, "Attention is all you need," in *Advances in Neural Information Processing systems*, 2017, pp. 5998–6008.

[13] J. Fu, J. Liu, H. Tian, Y. Li, Y. Bao, Z. Fang, and H. Lu, "Dual attention network for scene segmentation," in *Proceedings of the IEEE Conference on Computer Vision and Pattern Recognition*, June 2019.

[14] H. Zhang, I. Goodfellow, D. Metaxas, and A. Odena, "Self-attention generative adversarial networks," in *Proceedings of the 36th International Conference on Machine Learning*, ser. Proceedings of Machine Learning Research, K. Chaudhuri and R. Salakhutdinov, Eds., vol. 97. PMLR, 09–15 Jun 2019, pp. 7354–7363.

[15] S. Woo, J. Park, J.-Y. Lee, and I. S. Kweon, "Cbam: Convolutional block attention module," in *Proceedings of the European Conference on Computer Vision*, September 2018.

[16] K. He, X. Zhang, S. Ren, and J. Sun, "Identity mappings in deep residual networks," in *Proceedings of the European Conference on Computer Vision*, B. Leibe, J. Matas, N. Sebe, and M. Welling, Eds. Cham: Springer International Publishing, pp. 630–645.

[17] B. Zhou, A. Khosla, A. Lapedriza, A. Oliva, and A. Torralba, "Learning deep features for discriminative localization," in *Proceedings of the IEEE Conference on Computer Vision and Pattern Recognition*, 2016, pp. 2921–2929.